\documentclass{article}

\usepackage{graphicx} 
\usepackage{subfigure} 

\usepackage{natbib}

\usepackage{algorithm}
\usepackage{algorithmic}

\usepackage{hyperref}

\usepackage[accepted]{icml2013}

\icmltitlerunning{Learning Functions in Large Networks requires Modularity and produces Multi-Agent Dynamics}

\begin{document} 

\twocolumn[
\icmltitle{Learning Functions in Large Networks requires Modularity and produces Multi-Agent Dynamics}

\icmlauthor{C.-H. Huck Yang*}{chao-han.yang@kaust.edu.sa}
\icmladdress{Living Systems Laboratory, BESE, CEMSE, King Abdullah University of Science and Technology, Thuwal, KSA* \\ Georgia Institute of Technology, North Ave NW, Atlanta, USA. }
\icmlauthor{Rise Ooi*}{rise@eis.hokudai.ac.jp}
\icmladdress{Hokkaido University, Sapporo, Hokkaido, Japan.}
\icmlauthor{Tom Hiscock}{twh27@cam.ac.uk}
\icmladdress{University of Cambridge, UK}
\icmlauthor{Victor Eguiluz}{victor@ifisc.uib-csic.es}
\icmladdress{IFISC (CSIC-UIB), Spain}
\icmlauthor{Jesper Tegner}{jesper.tegner@kaust.edu.sa}
\icmladdress{Living Systems Laboratory, BESE, CEMSE, King Abdullah University of Science and Technology, Thuwal, KSA}

\icmlkeywords{gene circuits, machine learning, synthetic biology}

\vskip 0.3in
]

\begin{abstract} 

Networks are abundant in biological systems. Small sized over-represented network motifs have been discovered, and it has been suggested that these constitute functional building blocks. We ask whether larger dynamical network motifs exist in biological networks, thus contributing to the higher-order organization of a network. To end this, we introduce a gradient descent machine learning (ML) approach and genetic algorithms to learn larger functional motifs in contrast to an (unfeasible) exhaustive search. We use the French Flag (FF) and Switch functional motif as case studies motivated from biology. While our algorithm successfully learns large functional motifs, we identify a threshold size of approximately 20 nodes beyond which learning breaks down. Therefore we investigate the stability of the motifs. We find that the size of the real negative eigenvalues of the Jacobian decreases with increasing system size, thus conferring instability. Finally, without imposing learning an input-output for all the components of the network, we observe that unconstrained middle components of the network still learn the desired function, a form of homogeneous team learning. 
We conclude that the size limitation of learnability, most likely due to stability constraints, impose a definite requirement for modularity in networked systems while enabling team learning within unconstrained parts of the module. Thus, the observation that community structures and modularity are abundant in biological networks could be accounted for by a computational compositional network structure.

\end{abstract} 

\section{Introduction}

The structure of networks has been investigated extensively using global metrics such as distributions of links and nodes, and local structures such as network motifs and graphlets. Learning algorithms, on the other hand, produce for example large deep layered structures with powerful pattern detection capabilities. In contrast, designed engineered systems are as a rule built from dynamical computational motifs or gates, producing a specific input-output function. However, due to combinatorial complexity only up to small three node functional motifs have been investigated.
Network motifs~\cite{milo2002network} are the local features which are statistically over represented in networks. Investigators have designed computational searching methods to detect of over-represented motifs~\cite{cussat2011artificial}. However, statistical approaches include Monte Carlo methods and Bayesian models~\cite{xing2004motifprototyper} are still computationally exhaustive and rely on existing quantitative data. GeneNet~\cite{hiscock2017adapting} is a developed neural network algorithm for gene circuit design without requiring prior data, which reinforce random initial networks to learn desired output functions by computing arbitrary functions (e.g., detect vibration of specified duration) and manipulating motifs behaviors (e.g., count pulses) in short computing time (e.g., $28.5$ sec.) The previous results from GeneNet has shown its relevance on existing FF and Switch motifs studies (e.g., the classical FF motif from \cite{cotterell2010atlas}) In this paper, we further investigate the connectivity and stability of GeneNet on high-ordered network components, as from $3\times3$ to $18\times18$ matrix. Through the regularization, genetic searching and stability mapping, we verified the reproducibility of this method as generative models on learnable networks as a functional multi-agent learning process. In contrast to exhaustive search, our approach is not limited to the identification of only small network motifs but more on the stability analysis with the high-ordered functional bio-motifs. 

Functional gene circuit such as FF \cite{cotterell2010atlas} and Switch \cite{gardner2000construction} have been further investigated on the modeling to analysis on $in$ $vitro$. The previous \cite{joachimczak2009evolution,chavoya2008cell} works on functional gene circuit design have been widely explored the FF model on three dimensional by physical simulation and genetic algorithm on a dynamic system. However, those approaches demand larger computing time and barely cover the feasibility of algorithm-based high-ordered gene circuit design. 

Higher-ordered structures in networks \cite{lu2009next} have become increasingly important since its applications and potential on understanding dynamically probing biological systems \cite{mutalik2013precise}. Engineering of artificial gene networks \cite{ellis2009diversity} has used control engineering principles to guide predictable gene network construction $in $  $silico$. However, these engineered transcriptions ~\cite{khalil2010synthetic} factors have not yet been fully characterized, and if they are to be used as building blocks for complex gene networks. Instead of knowledge-based exclusive searching models, recent progress on ML-based techniques provides available software, etc. PyTorch and Tensorflow, tools to develop learning-based algorithms \cite{bengio2013representation} to describe and analyze the bio-system ~\cite{grover2016node2vec} in a high-ordered network. learning-based method for high-ordered bio-network would serve to investigate deeper on the $in$ $vivo$ kinetics and input-output transfer functions for synthetic biology~\cite{wu2016metabolic}.

\begin{figure}
\begin{center}
   \includegraphics[width=1.0\linewidth]{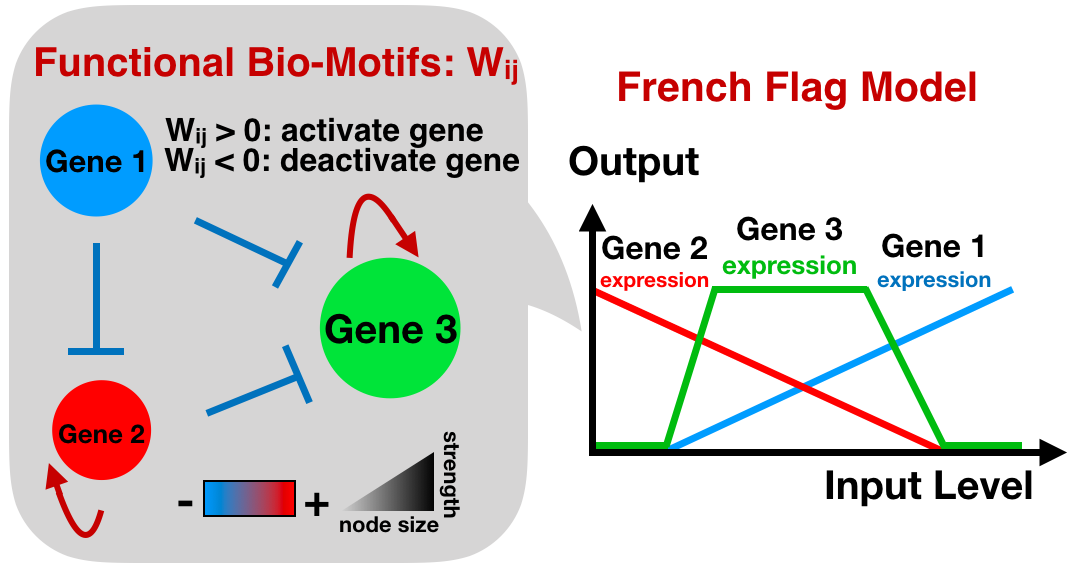}
\end{center}
   \caption{An example of a gene-circuit motifs which computes the French-Flag input-output function. The interactions between the units define the adjacency matrix.}
\label{fig:figure1}
\end{figure}

\textbf{Contributions.}
\begin{itemize}
    \item Based on existing ODE-based motifs learning method, we further expand the ML-based gene circuit design in a large network and propose a modified version of GeneNet, ES-GeneNet by evolutionary searching.
    \item We conduct adaptation behaviors on both homogeneous and heterogeneous team-learning among different sizes of a network. 
    \item Finally, we find out the two specific network behaviors on size effects and evolving dynamics by the addressing stability analysis of multi-agent learning in bio-system.
\end{itemize}

\begin{figure*}
\begin{center}
   \includegraphics[width=1.0\linewidth]{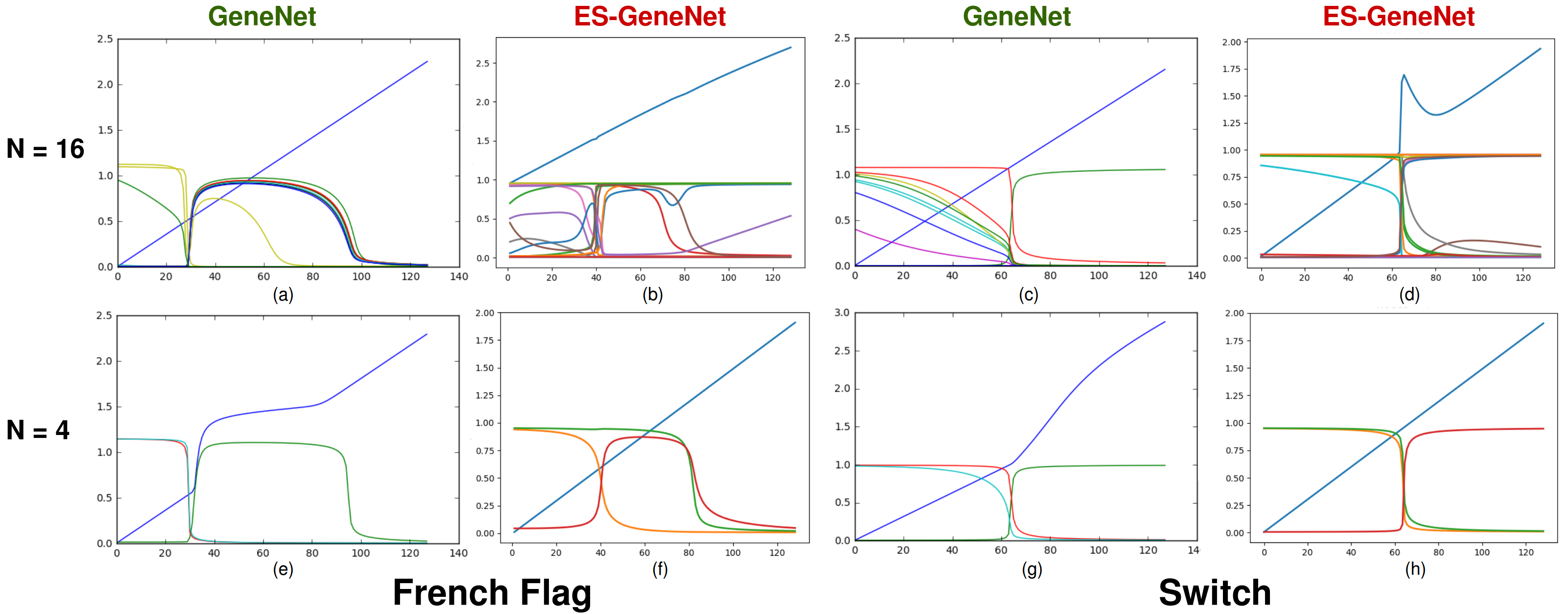}
\end{center}
   \caption{Learning larger network motifs. Both gradient descent (GeneNet) and genetic algorithms (ES-GeneNet) can learn functional network motifs larger than 3 nodes as discovered previously. Two examples with $4$ and $16$ nodes are illustrated for the cases of FF and the switch function. At a larger size of $16$, the learnable gene circuits perform for FF on (a) and (b) and Switch on (c) and (d). For the smaller size of $4$, the learnable gene circuits perform for FF on (e) and (f) and Switch on (g) and (h).}
\label{fig:figure2}
\end{figure*}

\section{Methods}
\subsection{Model of Gene Circuits}

Dynamic gene network models \cite{reil1999dynamics} describe the time-dependent evolution of gene expression. Each node in the network receives input from the other nodes \cite{de2002modeling}. We follow the formulation of the gene network in \cite{hiscock2017adapting, cotterell2010atlas}:  \\
\[
\frac{dy_i}{dt} = \phi(\sum_{i,j}{W_{ij}y_{j}})+I_i-y_i ~~~~(1)
\]

  where the $y_i$ denotes the concentrations of $i$th gene, where $i=1 ... N$ for an N-node network. $W_{ij}$ is a matrix, $W\in R^{n \times n}$, correspond to network weight: $W_{ij}>0$ means that $i$-th activate the $j$-th gene, and $W_{ij}<0$ means that $i$-th deactivate the $j$-th gene. For manipulating desired input and the output function, $y_i = F(x)$ , where we define $y_1 \equiv x$ and $y_n \equiv y$ (for output at the final node.) $I_i$ is the prescribed impulse, we give to the gene-network system. The $\phi(x)$ is a nonlinear function, ensuring that transcription rate to be activated. We choose the sigmoid function as $\phi(x)$ for activating neuron computing unit as
 \[
\phi(x)=\frac{1}{1+e^{-x}} ~~~~~~~~~~~~~(2)
\]

Genes expression and logic components could describe the initial condition (weight) of the first node and final output of the last node. Previous work \cite{hiscock2017adapting} of GeneNet focused on 3-nodes system, which automatically learn the desired function to design a proper network computing the prescribed input-output function. We use this approach to train models for the French-Flag (FF) \cite{miller2004evolving} and a switch function.  

\subsection{Learning parameters using Gradient Descent}
The ordinary differential equation (ODE) model of GeneNet in the equation $(1)$ shares similarities with the underlying neural network model on the matrix computing by considering $I_i$ (prescribed impulse) as bias $b_i$. The basic form of the simplified classical neural network would be: \\
\[
y_i = \phi(\sum_{i,j}{W_{ij}x_{j}}+b_i) ~~~~~~~(3)
\]


By giving desired output function of the network at $\hat{y}$, such as FF model, we minimize cost between the training output and the desired function by Adam \cite{kingma2014adam}, an effective gradient descent algorithm for this problem. In this paper, we use mean squared error as the cost function lessened by gradient descent in TensorFlow 1.7 module. The mean squared error (MSE) cost function has been defined as:
\[
\mathop{\arg\min}_{x} ||(y(x)-\hat{y})||_{2}^{2} ~~~~~~~(4)
\]

We use the GeneNet framework with the modified learning rate for the higher-ordered network experiments in this paper. In contrast to the current BP gradient descent technique as implemented in GeneNet, our proposed ES-GeneNet, an evolutionary optimization for cost minimization would be discussed in $section$ $3.3$.  

\subsection{LASSO Regularization}
To avoid over-fitting on the generated solutions, LASSO (least absolute shrinkage and selection operator, LASSO) has been used as a regression analysis method that performs both variable selection and regularization ~\cite{lim2015learning} to enhance the prediction accuracy and interpretability of the statistical model it produces. We modify the existing MSE cost function as :
\[
Cost(x)=\frac{1}{n}||y(x)-\hat{y}||^{2}_{2}+\lambda\sum_{i,j}{||W_{ij}||_{1}} ~~~~(5)\\
\]

With different L1 coefficients, we calculate the strength of genes by $\sum_{i,j} \left | W_{i,j}  \right |$. Where $\lambda$ represents the regularization coefficients, after testifying the the value among 
$ {2\times10^{-3}}-{2\times10^{-1}}$ and the under-fitting occur at ${2\times10^{-1}}$ with an failure case on desired functional motifs. We use L1 coefficients ${2\times10^{-2}}$ for the regularized GeneNet experiments in Figure \ref{fig:figure3}, \ref{fig:figure4}, \ref{fig:figure5}, and \ref{fig:figure6}.

\begin{table}[]
\centering
\caption{L1 Regularization with different coefficients}
\label{my-label}
\begin{tabular}{lllll}
\hline
Coeff.   & $0$ & $2\times10^{-1}$ & $2\times10^{-2} $ & $2\times10^{-3}$ \\ \hline
Strength & -2.51  & 0.001             & 0.34              & 0.02             \\ \hline
\end{tabular}
\end{table}

\subsection{Learning parameters using an Genetic Algorithm}
Using gradient descent only a limited number of networks motifs is discovered that computed the FF function. We, therefore, explored whether an evolutionary approach finds other solutions compared to a gradient descent search, i.e., Equation $(1)$. 

An evolutionary algorithm ~\cite{fonseca1993genetic} generate many diverse solutions among global searching domain. By relying on bio-inspired operators such as mutation, crossover, and selection, the modified weight from ES-GeneNet has more considerable diversity compared to the single pathway based on gradient descent from GeneNet. We implement the evolutionary algorithms to replace each iteration of the updating gradient to generations. We arrange 20 different updating gradients, like children in an evolutionary algorithm for each generation. The time of computation is 2.22 times faster than the result from GeneNet at three-nodes components searching. 

\subsubsection{Crossover}
First, we use a crossover rate of $(60\%)$ to enable the different weights matrices to share their learned features during the process of cost minimization. The crossover action ensures the cost function to be convex in a large search space where the search of the desired weights correspond to the network motifs. 

\subsubsection{Mutation}
 Mutation is a genetic operator used to preserve genetic diversity from one generation of a population of evolutionary algorithm chromosomes to the next. We use mutation in two different ways in modified versions of GeneNet $(1)$. The gradient computation is performed but with a mutation rate $\frac{0.5}{N\times N}$ $(2)$ while doing without the backpropagation algorithm and secondly, to use random mutation and crossover to minimize the cost. 
 Algorithm~\ref{alg:example} shows an example. 

\begin{algorithm}[tb]
   \caption{Evolutionary Search}
   \label{alg:example}
\begin{algorithmic}
   \STATE {\bfseries Input:} $desired\_target\_function$ $Y$
   \STATE Initialize $population = randomized\_children()$.
   \FOR{$g=1$ {\bfseries to} $max\_possible\_generations$}
   \STATE $costs = calculate\_cost(population, Y)$
   \STATE $survived\_parents = select(population, costs)$
   \STATE $elite = survived\_parents[0]$
   \IF{$is\_desired\_function(elite, Y)$} 
   \STATE Break loop
   \ELSE
   \STATE $population = repopulate(survived\_parents)$
   \STATE $population = mutate(population)$
   \ENDIF
   \ENDFOR
   \STATE Output
\end{algorithmic}
\end{algorithm}

\begin{figure}
\begin{center}
   \includegraphics[width=1.0\linewidth]{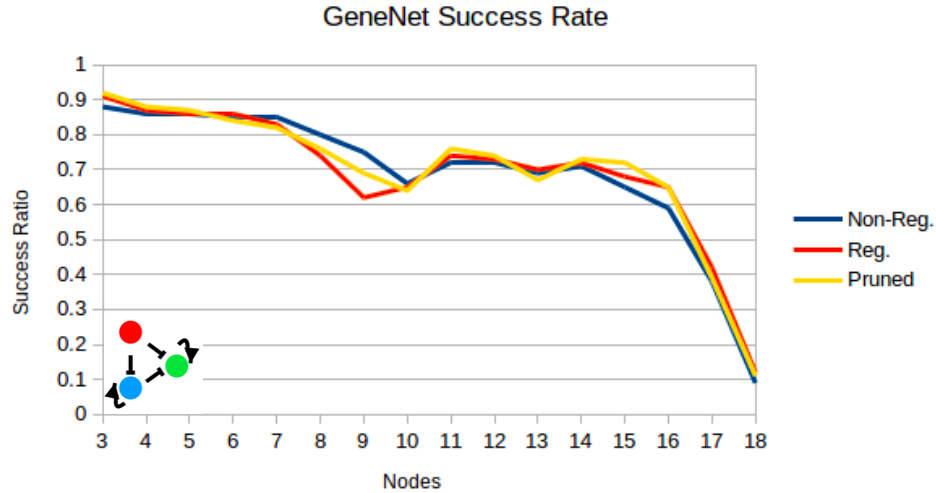}
\end{center}
   \caption{This figure illustrates the learnability of GeneNet on French-Flag function with increasing network size. The GeneNet algorithm is repeated 100 times and we calculate the ratio of successful learning of the FF function out of the 100 trials. Note the sharp drop around 16-18 nodes. This behaviour is consistent regardless of the degree of regularization (L1 or pruning, red and yellow respectively)}
   \vspace{-0.5cm}
\label{fig:figure3}
\end{figure}

\subsection{Numerical Implementation}
We run our code by Tensorflow 1.7.0 on a workstation with an Intel(R) Xeon(R) CPU E5-2680 v4 2.40GHz and a 32 GB Hynix DIMM Synchronous 2133 MHz. The code of ES-GeneNet is availble at ${github.com/huckiyang/ES}$\_ ${GeneNet}$.

\begin{figure*}
\begin{center}
   \includegraphics[width=0.95\linewidth]{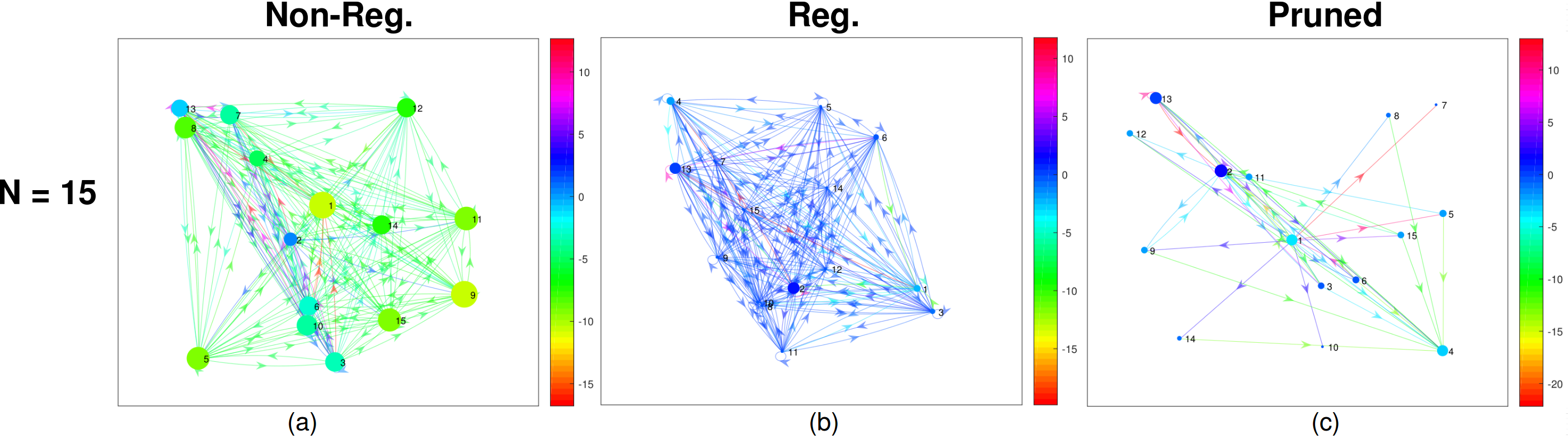}
\end{center}
   \caption{This figure visualizes the network interaction of learnable gene regulatory network in different gene strength on the LASSO (L1) regularization. At the network size of 7 (a) GeneNet without regularization, (b) GeneNet with L1 regularization, and (c) GeneNet with pruned L1 regularization.  }
\vspace{-0.5cm}
\label{fig:figure4}
\end{figure*}

 \section{Results and Discussion}
 
 \subsection{Identification of larger ($N>3$) Networks Motifs}
 First, we ask whether the gradient descent search and the genetic algorithm could find larger motifs to compute a given input-output function. The French-Flag (FF) model, is one of the most important and frequent models for gene-gene regulatory \cite{miller2004evolving}. By given desired function as $\hat{y}$ from the equation $(4)$, we search for FF circuits larger than three nodes. For example, we use $y=1$ for $x>0.5$  $x<1.5$ else $y=0$ as a FF gene expression Model. We discover both 16-genes and 4-genes circuit motifs computing the FF functions, as illustrated in Figure \ref{fig:figure2} $(a), (b), (e),$ and $(f)$. 
 
We also design a gene-circuit with functional Switch model. As a crucial regulatory network model, the switch is constructed from any two repressible promoters arranged in a mutually inhibitory network. This functional gene circuit is flipped between thermal induction and exhibits a nearly ideal controlled threshold or stable states using a transient chemical. This function is parametrized as $y=1$ for $x>1$ else $y=0$ and also here we find larger circuits, as illustrated by the 16-gene and 4-gene circuit motif of Figure \ref{fig:figure2} $(c), (d), (g),$ and $(h)$.

 Parametrizing the FF model, using $y=1$ for $x>1$ else $y=0$ as a FF gene expression Model in 4-genes and 16-genes circuit as illustrated in Figure \ref{fig:figure2} $(c)$, produced on average a speed-up of the computation by factors 2.73 and 7.73 respectively as compared to GeneNet. Also, we empower the Switch model, as $y=1$ for $x>1$ else $y=0$ as a FF gene expression Model in 4-genes and 16-genes circuit of Figure \ref{fig:figure2} $(d)$ with a average of 2.52 and 5.52 times faster than the computation of GeneNet.
 
In mathematics and computer science, connectivity ~\cite{harris1988combined} asks for the minimum number of elements (nodes or edges) that need to be removed to disconnect the remaining nodes from each other. We calculate the undirected edge-connectivity ~\cite{nagamochi1992computing} from the generate matrix size from $3\times3$ up to the largest size we learn from the random generate $18\times18$ in Table 3. With identical desired training function, the result showed the more extensive network is represented by specific regularization motifs network to present the dynamic performance. 

The nodes size of the network represents the strength in a gene regulatory circuit. Here we visualize nodes size by the value of $\sum_{i,j} \left | W_{i,j}  \right |$. With stronger strength, the node would have a bigger size, which implies a dominant gene in the regulatory network. With fixed size of learned gene regulatory network from the Figure \ref{fig:figure4}, the value of $\sum_{i,j} \left | W_{i,j}  \right |$ would decreasing as a much neat network interaction without trivial nodes. As a input-output system, we also visualize the sum of system feedback, $\sum_{i,j} W_{i,j} $, with a color bar. For instance, as Figure \ref{fig:figure4} $(a)$, the $1st$ node with a green and the $2nd$ node with a blue represent as a sum of feedback with positive and negative response in a gene regulatory network.

Moreover, we also feature the learned network as a tool to show the activate and deactivate relationship in an extensive system as gene regulatory network. As part of the result in Figure \ref{fig:figure4} (c), in a visualized pruned-network, we efficiently investigate the dynamics of both positive and negative impacts for each node as each gene. 

\subsection{There is a threshold of the size of the network motif computing an input-output function}

We asked whether there is any limitation on the size of network motifs computing an input-output function. To quantify this, we use the ratio between the number of successful trials computing the function and the total number of attempts to compute the input-output function. This experiment is performed both for the FF and switch function. Surprisingly, we find a sharp drop at around 18 nodes as illustrated in Figure \ref{fig:figure3}. Now, why did this drop occur? On the one hand, it is a consequence that the search space simply became too large at a certain point and that the algorithms were unable to identify appropriate solutions. Alternative the ratio of the proper solution decreases rapidly due to the exponential increase in search space. However, several observations suggest that this is not the reason. First, we observed the same threshold, regardless if we use a gradient descent search or a genetic algorithm. Secondly, the threshold is very similar for both the FF and the switch function. 
To further examine this we visualize the network at the transition from being able to compute the function over to the case when the network failed. Figure \ref{fig:figure5} illustrates that the sign of edges in the network gradually shift from being negative (a) to become positive (c). Hence, it appears that the negative feedback regulation in the network vanishes at the threshold. This suggests that the stability of the network is affected.

\begin{figure*}[h]
\begin{center}
   \includegraphics[width=0.95\linewidth]{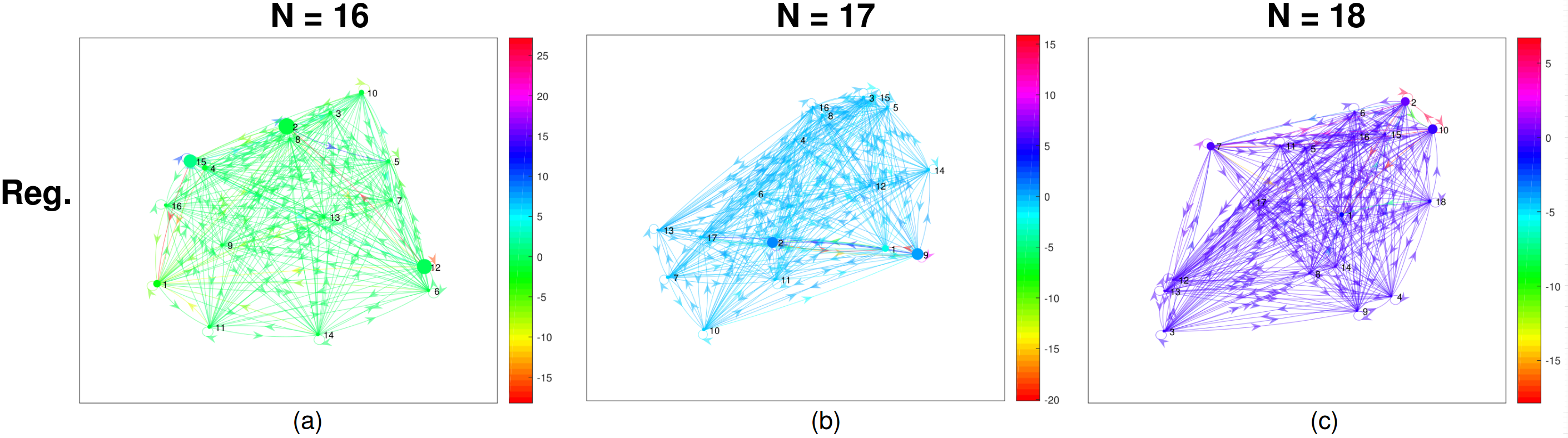}
\end{center}
   \caption{At the threshold of learnability there is a shift in the weight distribution from negative to positive weights. From a most negative weights at (a) nodes $= 16$, the shifting weights to near zero points at (b) nodes $= 17$, and finally most of the weights became positive at (c) nodes $= 18$. This shift in the weight distribution is in accordance with the stability map of eigenvalues as displayed in Figure \ref{fig:figure5}}.
\vspace{-0.5cm}
\label{fig:figure6}
\end{figure*}

\subsection{Stability analysis of the Network}

\begin{figure}[h]
\begin{center}
   \includegraphics[width=1.0\linewidth]{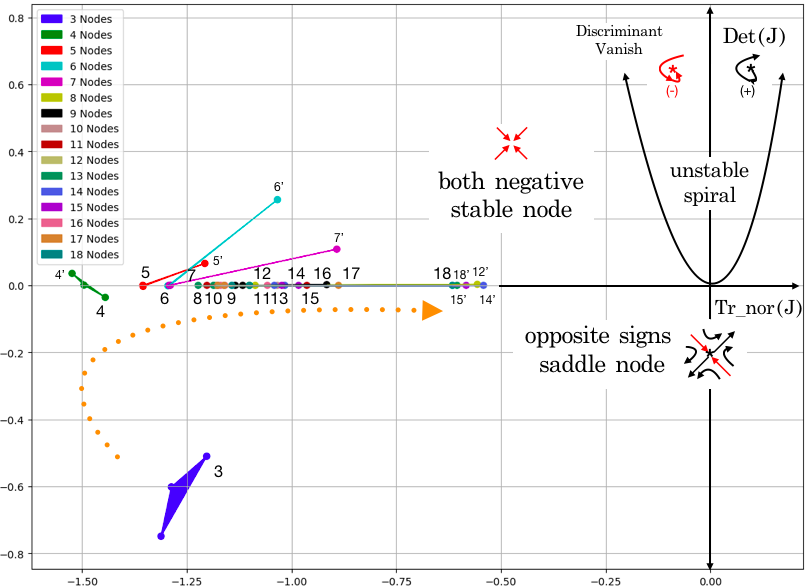}
\end{center}
   \caption{This figure show stability map of Jacobian matrices from bio-motifs. The process shows stability by routing different network size and evolutionary process on the map.  }
\label{fig:figure5}
\end{figure}

To investigate the stability we estimate the size of the negative real eigenvalues as a function of the network size. In general, a fixed-point for a network is stable if the real part of the eigenvalue is negative. Using $Equation (1)$, the adjacency matrix from the GeneNet has been activation by a non-linear equation. We take a Jacobian transformation of a learned adjacency matrix on the fixed point mapping. 
From the Figure \ref{fig:figure5}, as the orange arrow illustration, the following size effects gradually bring the stability from saddle points to stable point at the second quadrant, then at the size of 18 - the system closed to the zero-point as a threshold of stability. The regularized results among networks as $4'$, $5'$, $6'$ in a small network and $16'$, $17'$, $18'$ in a large network are echoed to the stability hypothesis on the fixed point theory.


\subsection{Adaptation on Multi-Agent Learning}
The previous study of GeneNet \cite{hiscock2017adapting} on learning functional motifs has succeeded in initializing random 3-nodes genes circuits. However, higher-ordered network learning on functional gene-circuits has not yet been explored using the neural network generative model. When analyzing larger learned network motifs opens up the question if and how the other elements in the system contributes to the learned input output function. This has been referred to as the team learning problem \cite{panait2005cooperative}. In brief, there is positive and negative feedback in such circumstances, which originates from the adaptation of the elements supporting team learning in order to compute the desired function. As our definition in Equation $(1)$, we only supervise the last node, as $y_n \equiv y$, and minimize its MSE during the learning processing. Interestingly, with an increasing size effect, we observed that the agents — nodes in our case — tend to perform two types of Team Learning behaviors.

\subsection{Homogeneous Team Learning}

Team learning is an easy approach to multi-agent learning because its single learned can use standard single-agent machine learning techniques. However, a major problem with team learning is the large state space for the learning process with large dynamic with two opposite behavior homogeneous and heterogeneous team learning. With systematic feedback results, we can observe the homogeneous learning behaviors on desired FF functional circuit for different size of the networks. Homogeneous Team-Learning is an effect of a multi-agent system tend to learn the same as given desired function, even without supervised. This homogeneous learning is positive for supporting our higher-ordered gene-circuit to learn band-pass performance in FF model. The amplitude of the targeted red FF signal, $y_n \equiv y$, slightly increase $5.1\%$ and $4.3\%$ from $(a)$ to $(b)$ and from $(c)$ to $(d)$. Meanwhile,  a series of un-supervised agents keep increasing their amplitude among the band-pass performing range.

\subsection{Heterogeneous Team Learning}

We find it to be different from previous observations as small as $9$. Negative feedback after the size attains a large size at 16 as a Heterogeneous Team Learning effect. And, the un-supervised agents are pulled back away from the desired features of the supervised agent. Meanwhile, our multi-agent network is gradually losing its learnability and abruptly crashed down at the size of $18\times18$ with a corresponding low successful learning rate at 9\% in \ref{fig:figure3}. The amplitude of the targeted red FF signal, $y_n \equiv y$, dramatically drop $12.1\%$ and $9.3\%$ from $(d)$ to $(e)$ and from $(e)$ to $(f)$.

\subsection{Summary of the observed Team Learning}

In summary, we found positive feedback from the un-supervised agents, i.e. the other nodes in the network motifs, echoes in their learned profiled the overall learned function in the network motif. Note, that this is unexpected as the supervised training is only conferred to the last node given the input to the first node, as the behaviour of all other nodes and interactions were left unconstrained. Note that the learned network motif input output function is only imposed on the system as a whole. This effect is functionally useful for learning in a complex system like the gene regulatory network. The homogeneous team learning turn to drag down the third agent from the band-pass performance and lead an unstable system. Also, note that the positive-feedback turn to opposite after the network size becomes too large. The un-supervised agent is drawn down the desired features from the supervised agent. Meanwhile, our multi-agent network is gradually losing its learnability and suddenly crash down at the size of $18\times18$ with a successful learning rate less than 9\%, see Figure \ref{fig:figure3}.


\begin{figure}[h]
\begin{center}
  \includegraphics[width=1.0\linewidth]{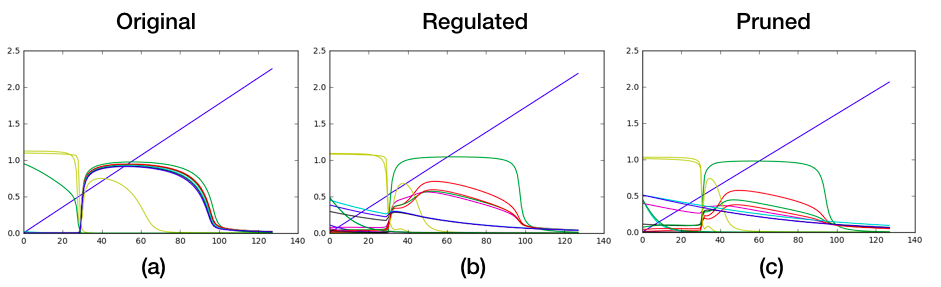}
\end{center}
  \caption{ This figure shows an changing system feedback from a homogeneous team learning to a heterogeneous team-learning effect at the network size/agent numbers = 17 with (a) original (b) regulated (c) pruned functional French-Flag gene-circuit }
\label{fig:figure7}
\end{figure}

\section{Conclusion}
In summary, we demonstrated the existence of larger functional input output network motifs in contrast of what have previously been described. To achieve this we had to adapt an ML approach to design and search for functional high-order gene regulatory network motifs. The rational is that an exhaustive search for larger network motifs does not scale due to the combinatorial complexity in the search space with increasing size of the network motifs. To this end we use both gradient descent and genetic algorithms. The observed threshold in learning that we observed is most likely due to loss of stability with increasing size. This effect requires further investigation as it would most likely have to with the specific structure of the Jacobian matrix as the system size grows. Hence, the elements which are not directly constrained in the learning, i.e. the nodes in between the input and output have to organize themselves in such a manner that stability is preserved for as large system as possible. However, for some reason this does not apparently scale indefinitely. We conjecture that this phenomena has to do with the observed adaptive learning $in$ $vivo$ multi-agent behavior that we observed in the network motifs.  
Our proposed novel evolutionary algorithm framework to solve the ODE system without being trapped by local minimum with a single phenotype of FF-motifs hold the promise to be used for massive large-scale explorations. Both crossover mutation actions increase the diversity of the search space to find more diverse motifs-weight compared to the previous gradient descent method. This opens the possibility to not only identify multiple motifs of a given size for an input-output function. It could also explore higher-ordered motifs with adaptive team-learning in dynamic bio-system. Thus, our further work will focus on investigating biological properties and the analyze the high frequent motifs in the learned dynamic networks from GeneNet and ES-GeneNet. 

\section*{Acknowledgement}

This work is supported by competitive research funding from King Abdullah University of Science and Technology (KAUST). Also, we would like to acknowledge Google Cloud Platform, Google Dubai.



\nocite{langley00}

\bibliography{example_paper}
\bibliographystyle{icml2013}

\end{document}